%% file: main.tex
\def\year{2020}\relax
\documentclass[letterpaper]{article} 
\usepackage{aaai20}  
\usepackage{times}  
\usepackage{helvet} 
\usepackage{courier}  
\usepackage[hyphens]{url}  
\usepackage{graphicx} 
\usepackage{graphicx}  
\usepackage{subfigure}
\usepackage{nicefrac}       
\usepackage{algorithm}
\usepackage{algorithmic}
\usepackage{dsfont}
\usepackage{MnSymbol}
\usepackage{mathtools}
\usepackage{amsmath,amsthm}
\usepackage{amsfonts}       

\newtheorem{prop}{Proposition}

\newtheorem{ex}{Example}
\newtheorem{asp}{Assumption}

\DeclarePairedDelimiter{\norm}{\lVert}{\rVert}
\DeclareMathOperator*{\argmax}{arg\,max}
\DeclareMathOperator*{\argmin}{arg\,min}

\newcommand{\E}{\mathbb{E}}
\newcommand{\R}{\mathbb{R}}

\DeclarePairedDelimiterX{\infdivx}[2]{(}{)}{%
  #1\;\delimsize|\delimsize|\;#2%
}

\newcommand\citealp[1]{\citeauthor{#1} (\citeyear{#1})}

\title{Cost-Effective Incentive Allocation via Structured Counterfactual Inference}
\author{
Romain Lopez\textsuperscript{\rm 1}~, Chenchen Li\textsuperscript{\rm 2, 3}, Xiang Yan\textsuperscript{\rm 2, 3}, Junwu Xiong\textsuperscript{\rm 2}, \\ \Large
\textbf{Michael I. Jordan\textsuperscript{\rm 1}, Yuan Qi\textsuperscript{\rm 2}, and Le Song\textsuperscript{\rm 2, 4}}\\
\textsuperscript{\rm 1}Department of Electrical Engineering and Computer Sciences, University of California, Berkeley\\
\{romain\_lopez, jordan\}@cs.berkeley.edu \\
\textsuperscript{\rm 2}AI Department, Ant Financial Service Group\\
\{junwu.xjw, yuan.qi, le.song\}@antfin.com
\textsuperscript{\rm 3}Department of Computer Science, Shanghai Jiao Tong University\\
lcc1992@sjtu.edu.cn, xyansjtu@163.com \\
\textsuperscript{\rm 4}College of Computing, Georgia Institute of Technology
}

\begin{document}

\maketitle

\input{0_abstract}
\input{1_introduction}

\input{2_related_work}

\input{3_problem_setup}

\input{4_theory_and_algs}

\input{5_experiments}

\input{6_discussion}

\bibliographystyle{aaai}
{\fontsize{9pt}{10pt} \selectfont \bibliography{biblio}}

\onecolumn
\input{supp_material}

\end{document}

%% file: 0_abstract.tex
\begin{abstract}
We address a practical problem ubiquitous in modern marketing campaigns, in which a central agent tries to learn a policy for allocating strategic financial incentives to customers and observes only bandit feedback. 
In contrast to traditional policy optimization frameworks, we take into account the additional reward structure and budget constraints common in this setting, and develop a new two-step method for solving this constrained counterfactual policy optimization problem. Our method first casts the reward estimation problem as a domain adaptation problem with supplementary structure, and then subsequently uses the estimators for optimizing the policy with constraints. We also establish theoretical error bounds for our estimation procedure and we empirically show that the approach leads to significant improvement on both synthetic and real datasets.
\end{abstract}

%% file: 1_introduction.tex
\section{Introduction}

\noindent Batch Learning from Bandit Feedback (BLFB)~\cite{Swaminathan2015CF,Swaminathan2015self} is a form of counterfactual inference given only observational data~\cite{Pearl}. The problem arises in many real-world decision-making scenarios, including personalized medicine, where one is interested in estimating which treatment would have led to the optimal outcome for a particular patient~\cite{Saria}, or online marketing, which might focus for example on placing ads to maximize the click-through-rate~\cite{Strehl2010}.

In this paper, we focus on a novel flavor of BLFB, which we refer to as \textit{cost-effective incentive allocation}. In this problem formulation we allocate economic incentives (e.g., online coupons) to customers and observe a response (e.g., whether the coupon is used or not). Each action is mapped to a cost and we further assume that the response is monotonically increasing with respect to the action's cost. Such an assumption is  natural in many of the problems domains that motivate us, including drug-response estimation and online marketing, and it has the virtuous side effect of improving generalization to test data and making the model more interpretable. We also incorporate budget constraints related to the global cost of the marketing campaign or treatment regime. This framework can be readily applied to the problem of allocating monetary values of coupons in a marketing campaign under fixed budget constraints from the management. 

Existing work in counterfactual inference using bandit feedback~\cite{Joachims2018,Shalit2016} does not make use of the supplementary structure for the rewards and is limited in practice when the cardinality of the action space is large~\cite{LefortierSGJR16}. We therefore developed a novel algorithm which incorporates such structure. The algorithm, which we refer to as \textit{Constrained Counterfactual Policy Optimization via Structured Incentive Response Estimation} (CCPOvSIRE), has two components. First, we build on recent advances in representation learning for counterfactual inference~\cite{Johansson2016} and extend this framework to estimate the incentive response for multiple treatments while taking into account the reward structure (SIRE). Second, we rely on the estimates to optimize the coupon assignment policy under budget constraints (CCPO).

The remainder of the paper is organized as follows. First, we review existing approaches related to the problem of cost-effective incentive allocation and provide a brief background on BFLB. Then, we present our novel setting and the corresponding assumptions. In particular, we introduce structure on the action space, cost functions and budget constraints. Furthermore, we derive theoretical upper bounds on the reward estimates and present CCPOvSIRE as a natural approach to optimizing this bound. Finally, we evaluate our approach on fully-simulated data, and introduce a novel benchmarking approach based on nested classification which we apply to semi-simulated data from ImageNet~\cite{ImageNet}. In both cases, we show that our method compares favorably to approaches based on BanditNet~\cite{Joachims2018}, CFRNet~\cite{Shalit2016}, BART~\cite{hill2011} and GANITE~\cite{yoon2018ganite}.

%% file: 2_related_work.tex
\section{Related Work}\label{sec:related_work}

\paragraph{Constrained Policy Optimization} Safety constraints in reinforcement learning (RL) are usually expressed via the level sets of a cost function. The main challenge of safe RL is that the cost of a certain policy must be evaluated with a off-policy strategy, which is a hard problem~\cite{jiang16}. Recent work focuses on developing a local policy search algorithm with guarantees of respecting cost constraints~\cite{Achiam2017}. This approach is based on a trust-region method~\cite{Schulman2015}, which allows it to circumvent the off-policy evaluation step. Our setting is more akin to that of contextual multi-armed bandits, since customers are modeled as independent replicates from a unique distribution. In this scenario, checking for the satisfaction of the cost constraints is straightforward, which makes the original problem substantially easier. Most research contributions on policy optimization with budget constraints focus on the online learning setting~\cite{ding2013multi}
~and are therefore not directly applicable to BLFB.

\paragraph{Counterfactual Risk Minimization (CRM)} The problem of BLBF consists of maximizing the expected reward using importance sampling (IS), as follows: 
\begin{align}
\label{eq:pop}
    \E_{x \sim \mathcal{P}(x)} \E_{y \sim \rho(y \mid x)} \left[\delta(x, y) \frac{\pi(y \mid x)}{\rho(y \mid x)} \right], 
\end{align}
where $\delta$ is the reward function, $\pi$ a parameterized policy and $\rho$ the logging policy. Notably, the action probabilities $\rho(y \mid x)$ need to be either logged or estimated. In the specific case where the logging policy is unknown and learned from the data, error bounds have been derived~\cite{Strehl2010}. The variance of the naive IS estimator can be reduced using a doubly robust estimator~\cite{Dudik2011}. The CRM principle~\cite{Swaminathan2015CF} is based on empirical Bernstein concentration bounds~\cite{Maurer2009} of finite-sample estimates for Eq.~\eqref{eq:pop}. 
BanditNet~\cite{Joachims2018} is a deep learning methods based on equivariant estimation~\cite{Swaminathan2015self} and stochastic optimization. An advantage of this framework is that its mathematical assumptions are weak, while a disadvantage is that it is not clear how to make use of structured rewards.

\paragraph{Estimation of Individualized Treatment Effect (ITE)} The ITE~\cite{pmlr-v80-alaa18a} is defined as the difference in expectation between two treatments 
\begin{align}
\label{eq:ITE}
    \E[r \mid x, y=1]  - \E[r \mid x, y=0],
\end{align}
where $x$ is a point in customer feature space and $r$ is a random variable corresponding to the reward. The difficulty of estimating the ITE arises primarily from the fact that historical data do not always fall into the ideal setting of a randomized control trial. That inevitably induces an estimation bias due to the discrepancy between the distributions $\mathcal{P}(x \mid y = 0)$ and $\mathcal{P}(x \mid y = 1)$. BART~\cite{hill2011} 
is an estimation procedure based on Bayesian nonparametric methods. CFRNet~\cite{Shalit2016} and BNN~\cite{Johansson2016} both cast the counterfactual question into a domain adaptation problem that can be solved via representation learning. Essentially, they propose to find an intermediate feature space $\mathcal{Z}$ which embeds the customers and trades off the treatment discrepancy for the reward predictability. GANITE~\cite{yoon2018ganite} proposes to learn the counterfactual rewards using generative adversarial networks and extends this framework to the multiple treatment case via a mean-square error loss. Notably, this line of work does not require knowing the logging policy beforehand. Remarkably, out of all these contributions, only one focuses on the setting of structured rewards and only in the case of binary outcomes~\cite{Kallus2019}.

%% file: 3_problem_setup.tex
\section{Background: Batch Learning from Bandit Feedback}
For concreteness, we focus on the example of a marketing campaign. 
Let $\mathcal{X}$ be an abstract space and $\mathcal{P}(x)$ a probability distribution on $\mathcal{X}$. We consider a central agent and let $(x_1, \ldots, x_n) \in \mathcal{X}^n $ denote a set of customers. We assume that each customer is an independently drawn sample from $\mathcal{P}(x)$. Let $\mathcal{Y}$ be the set of financial incentives which can be provided by the central agent and let $\mathcal{S}^\mathcal{Y}$ be the space of probability distributions over $\mathcal{Y}$. The central agent deploys a marketing campaign which we model as a policy $\pi: \mathcal{X} \rightarrow \mathcal{S}^\mathcal{Y}$. For simplicity, we also denote the probability of an action $y$ under a policy $\pi$ for a customer $x$ using conditional distributions $\pi(y \mid x)$. In response, customers can either choose to purchase the product from the central agent or from another unobserved party. Given a context $x$ and an action $y$, we observe a stochastic reward $r \sim p(r \mid x, y)$. In practice, the reward can be defined as any available proxy of the central agent's profit (e.g., whether the coupon was used or not). Given this setup, the central agent seeks an optimal policy:
\begin{align}
\pi^* \in \argmax_{\pi \in \Pi} \E_{x \sim \mathcal{P}(x)} \E_{y \sim\pi(y \mid x)} \E [r \mid x, y].
\label{ERM}
\end{align} 
This problem, referred to as BLBF, has connections to causal and particularly counterfactual inference. As described in~\citealp{Swaminathan2015CF}, the data are incomplete in the sense that we do not observe what would have been the reward if another action was taken. Furthermore, we cannot play the policy $\pi$ in real time; we instead only observe data sampled from a logging policy $\rho(y \mid x)$. Therefore, the collected data are also biased since actions taken by the logging policy $\rho$ are over-represented. 

\section{Cost-Effective Incentive Allocation}
\label{sec:setup_new}
The BLBF setting might not be suitable when actions can be mapped to monetary values, as we illustrate in the following example.
\begin{ex} \label{ex:MV}\emph{Monetary marketing campaign}. Let $y$ denote the discount rate and $x$ denote a customer's profile. Let $r$ be the customer's response (e.g., how much he or she bought). Since the customer will be more susceptible to use the discount as its monetary value increases, we assume that for each value of $x$, $r \mid x, y$ is almost surely increasing in $y$. Then, the trivial policy which always selects the most expensive action is a solution to problem \eqref{ERM}.
\end{ex}
This simple example motivates the main set of hypothesis we now formally introduce in order to better pose our problem. Notably, our added assumptions further relate actions to the reward distribution. 
\begin{asp} \emph{Structured action space and rewards}.
\label{ass:struct}
Let us note the conditional reward $\E[r \mid x, y]$ as $f(x, y)$. We assume there exists a total ordering $\prec_\mathcal{Y}$ over the space $\mathcal{Y}$. The reward distribution is compatible with the total ordering $\prec_\mathcal{Y}$ in the sense that
\begin{equation}
\begin{split}
\forall (x, y, y') \in \mathcal{X} \times \mathcal{Y}^2, y \prec_\mathcal{Y} y' \Rightarrow f(x, y) \leq f(x, y').
\end{split}
\end{equation}
\end{asp}
\begin{asp} \emph{Budget constraints}.
There exists a cost function $c: \mathcal{Y} \rightarrow \mathbb{R}^+$ monotone on $\mathcal{Y}$ with respect to the total ordering $\prec_\mathcal{Y}$. Let $m$ be a maximal average budget per customer. We define the set of feasible policies $\Pi$ as 
\begin{align}
\Pi = \{ \pi: \mathcal{X} \rightarrow \mathcal{S}^\mathcal{Y} \mid \E_{x \sim \mathcal{P}(x)} \E_{y \sim\pi(y \mid x)} [c(y)] \leq m \} .
\end{align}
\end{asp}
In this manuscript, we will assume that the ordering as well as the cost function are known. We subsequently tailor the BLFB problem to the constrained case as follows:
\begin{equation}
\begin{split}
 \max_{\pi \in \Pi} ~   & \E_{x \sim \mathcal{P}(x)} \E_{y \sim\pi(y \mid x)} f(x, y),  
\end{split} 
\label{CERM}
\end{equation}
which we refer to as \textit{Counterfactual Constrained Policy Optimization} (CCPO). Additionally, we refer to the problem of estimating the response function $f$ as \textit{Incentive Response Estimation} (IRE); this is a natural extension of the ITE problem in the scenario of multiple treatments as discussed in~\citealp{yoon2018ganite}. We now claim that IRE is a statistically harder problem than CCPO in the following sense:
\begin{prop}
\label{prop:eq}
Let us assume that the incentive response function $f$ is identified. Then solving Eq.~\eqref{CERM} for any budget $m$ reduces to a binary search over a unique Lagrange multiplier for the cost constraint. In particular, each step of the search has a complexity that is linear in the sample size. 
\end{prop}
\begin{proof}
See Appendix~\ref{app:proofs}.
\end{proof}
This negative results implies that, in general, plugging in the output of an estimation (IRE) algorithm for policy optimization (CCPO) might be suboptimal compared to directly learning a policy (the aim of CRM-based methods). However, Assumption~\ref{ass:struct} may substantially reduce the complexity of the estimation problem provided the inference procedure benefits from such structure (referred to as SIRE). As a consequence, we propose an approach that we refer to as ``CCPOvSIRE.'' We expect that such a structured counterfactual inference algorithm will outperform both ITE and CRM-based methods. This should especially be true in the regime of medium to large action spaces, where CRM-based methods struggle in practice~\cite{LefortierSGJR16}. Such tradeoffs between complexity and structure are common in machine learning (cf.\ discriminative versus generative approaches to supervised learning problems~\cite{NIPS2001_2020} and model-free versus model-based approaches to RL~\cite{pong2018temporal}).


%% file: 4_theory_and_algs.tex
\section{Constrained Counterfactual Policy Optimization via Structured Incentive Response Estimation}\label{sec:cpo_via_sire}
We adopt the following classical assumptions from counterfactual inference~\cite{rubin2005} which are sufficient to ensure that the causal effect is identifiable from historical data~\cite{Shalit2016}. Such assumptions explicitly imply that all the factors determining which actions were taken are observed. Notably, these must come from domain-specific knowledge and cannot be inferred from data.
\begin{asp} \label{ass:overlap}\emph{Overlap}. There exists a scalar $\epsilon > 0$ such that $\forall (x, y) \in \mathcal{X}\times\mathcal{Y}, \rho(y \mid x) > \epsilon$.
\end{asp}
\begin{asp} \emph{No unmeasured confounding}. Let $r^\mathcal{Y} = (r_y)_{y \in \mathcal{Y}}$ denote the vector of possible outcomes in the Rubin-Neyman potential
outcomes framework~\cite{rubin2005}. We assume that the vector $r^\mathcal{Y}$ is independent of the action $y$ given~$x$.
\end{asp}

We now turn to the problem of estimating the function $f$ from historical data, using domain adaptation learning bounds. To this end, we first write the estimation problem with a general population loss. Let $L: \R^2 \rightarrow \R+$ be a loss function and let $\mathcal{D}$ a probability distribution on the product $\mathcal{X}\times\mathcal{Y}$ (which we refer to as a \textit{domain}). As in \citealp{Shalit2016}, we introduce an abstract feature space $\mathcal{Z}$ and an invertible mapping $\Lambda$ such that $z = \Lambda(x)$. For technical developments, we need to assume that
\begin{asp}
$\Lambda: \mathcal{X} \rightarrow \mathcal{Z}$ is a twice-differentiable one-to-one mapping.
\end{asp}
This allows us to identify each domain $\mathcal{D}$ with a corresponding domain on $\mathcal{Z} \times \mathcal{Y}$. Let $(\hat{g}_\psi)_{\psi \in \Psi}$ be a parameterized family of real-valued functions defined on $\mathcal{Z}\times\mathcal{Y}$. We define the \textit{domain-dependent} population risk $\epsilon_\mathcal{D}\left(\hat{g}_\psi\right)$ as
\begin{align}
\epsilon_\mathcal{D}\left(\hat{g}_\psi\right) = \E_{(x, y) \sim \mathcal{D}} \left[ L\left(\hat{g}_\psi(\Lambda(x), y), f(x, y)\right) \right].
\label{IE}
\end{align} 
In the historical data, individual data points $(x, y)$ are sampled from the so-called \textit{source} domain $\mathcal{D}_S = \mathcal{P}(x)\rho(y \mid x)$. However, in order to perform off-policy evaluation of a given policy $\pi$, we would ideally need data from $\mathcal{P}(x)\pi(y \mid x)$. The discrepancy between the two distributions will cause a strong bias in off-policy evaluation which can be quantified via learning bounds from domain adaptation~\cite{Blitzer2007}. In particular, we wish to bound how much an estimate of $f$ based on data from the source domain $\mathcal{D}_S$ can generalize to an abstract target domain $\mathcal{D}_T$ (yet to be defined).

\begin{prop}
Let $\psi \in \Psi$, and $\mathcal{D}_S,\mathcal{D}_T$ be two domains. Let $\mathcal{H}$ be a function class. Let $\textrm{IPM}_\mathcal{H}\left(\mathcal{D}_S, \mathcal{D}_T\right)$ denote the \emph{integral probability metric} (IPM) between distributions $\mathcal{D}_S$ and $\mathcal{D}_T$ with function class $\mathcal{H}$
\begin{align}
    \textrm{IPM}_\mathcal{H}\left(\mathcal{D}_S, \mathcal{D}_T\right) = \sup_{h \in \mathcal{H}} \left| \E_{\mathcal{D}_S}h(x, y) - \E_{\mathcal{D}_T}h(x, y)\right|,
\end{align}
In the case that $\mathcal{H}$ is the set of test functions
 \begin{align*}
    \mathcal{H} = \left\{ (x, y) \mapsto L\left(\hat{g}_\psi(\Lambda(x), y), f(x, y)\right) \mid \psi \in \Psi \right\},
\end{align*} 
then, it is possible to bound the population risk on the target domain by
\begin{align}
\begin{split}
   \epsilon_{\mathcal{D}_T}\left(\hat{g}_\psi\right) \leq &~ \epsilon_{\mathcal{D}_S}\left(\hat{g}_\psi\right) + \textrm{IPM}_\mathcal{H}\left(\mathcal{D}_S, \mathcal{D}_T\right). \label{IPM}
\end{split}
\end{align}
\end{prop}
\begin{proof}
See Theorem 1 of~\citealp{Blitzer2007}.
\end{proof}
This result explicitly bounds the population risk on the target domain based on the risk on the source domain and the additional complexity term (IPM). We now explain how to compute both terms in Eq.~\eqref{IPM} and detail how we can make use of the structure from both theoretical and implementation perspectives.

\subsection{Source domain risk minimization}
From the algorithmic perspective, we can directly make use of the historical data to approximate the population risk on the source domain. In order to further benefit from the known structure about the reward function, we need to restrain the search space for $g_\psi$ to functions that are increasing in $y$ for each fixed transformed feature $z$. This problem has been studied in neural network architecture engineering~(MIN-MAX networks)~\cite{5443743}, and extensions are available to of architectures such as DLNs~\cite{NIPS2017_6891} or ICNNs~\cite{ICNN} could also be useful for other types of shape constraints (e.g., convexity). In our implementation, we rely on enforcing positivity on the weights of the last hidden layer as for the MIN-MAX networks. 

From an approximation theory standpoint, Eq.~\eqref{IPM} only refers to the population risk. However, benefits from the structure are expected in finite sample prediction error~\cite{wainwright_2019}, which can be linked back to population risk using uniform law arguments. Because analyzing prediction errors with neural networks may be a hard problem, we rely on well-studied nonparametric least-square estimators. However, while most research focuses on extending unidimensional results to multivariate function classes with the same assumptions (i.e., convexity in~\citealp{adityaconvex}, monotonicity of functions in~\citealp{gao2007entropy} and matrices in~\citealp{chatterjee2018}), asymmetry in the regularity assumptions between the different dimensions of a function class (such as partial monotony) has not received as much attention. We therefore derive such learning bounds in Appendix~\ref{app:nonparam}. In particular, we exhibit a better rate with respect to the sample size for a multivariate Sobolev space (continuous action space), under smoother assumptions. In the case of discrete actions $\mathcal{Y} = \{1, \ldots, K\}$, we use empirical processes theory to improve the dependence with respect to $K$. Namely, we investigate the case of functions with null variation, for which we present the result here.
\begin{prop}
    Let $\mathcal{F}$ denotes all functions defined on $[0, 1]\times \{1, \ldots, K\}$ and taking values in $[0, 1]$. Let us consider the following function classes
    \begin{align*}
    \begin{split}
        \mathcal{F}_c &= \left\{f \in \mathcal{F} \mid \exists a \in [0, 1]^K: \forall y : f(., y) = a_y \right\} \\
        \mathcal{F}_m &= \left\{f \in \mathcal{F}_c  \mid \forall x \in [0, 1], f(x, .) \text{~increasing} \right\}
    \end{split}
    \end{align*}
    Let $f^*$ in $\mathcal{F}_m$ be the incentive response, let $n \in \mathbb{N}$, $\sigma > 0$ and let us assume the noise model 
\begin{align*}
    \forall (i, j) \in \{1, \ldots, n\} \times \{1, \ldots, K\}, r_{ij} = f^*(\nicefrac{i}{n}, j) + \sigma w_{ij}, 
\end{align*}
with $w_{ij} \sim \text{Normal}(0, 1)$. Then with high probability, the least squares estimate $\hat{f}_m$ over $\mathcal{F}_m$ satisfies the bound 
\begin{align*}
\norm{\hat{f}_m - f^*}^2_n \leq \frac{\sigma^2\log K}{nK},
\end{align*}
where $\norm{.}_n$ denotes the empirical norm. Conversely, it is known that the least squares estimate $\hat{f}_c$ over $\mathcal{F}_c$ satisfies with high probability the bound 
\begin{align*}
\norm{\hat{f}_c - f^*}^2_n \leq \frac{\sigma^2}{n}.
\end{align*}
\end{prop}
\begin{proof}
See Appendix~\ref{app:nonparam}.
\end{proof}
Since our observation model here assume we observe for each $n$ the outcome of all the actions, we artificially inflated $n = n'K$. Intuitively, by treating this bound as a function of $n'$, we recover for $\mathcal{F}_c$ the expected parametric rate with a linear dependency in $K$. Such a rate is known to be minimax optimal (meaning the upper bound is indeed tight). Therefore, we proved that adding the structure on the rewards improves the dependency from $K$ to $\log K$, which is significant. Quantifying the benefit of structured rewards for more complex function classes is left as an open problem.  

\subsection{IPM estimation via kernel measures of dependency}
Thanks to the added structure, let us notice that the function class over which the supremum is taken inside the IPM of Eq.~\eqref{IPM} may be substantially smaller (i.e., restricted to partially monotonic functions). This implies that weaker regularization or weaker kernels may provide satisfactory results in practice. Such a result is also suggested by Theorem 1 of \citealp{pmlr-v80-alaa18a}. However, more systematic quantification of such an improvement is still an open problem. 

That said, Eq.~\eqref{IPM} assumes that we have at our disposal a \textit{target} domain $\mathcal{D}_T$ from which the estimated incentive response $\hat{g}_\psi$ would generalize to all policies for offline evaluation. We now explain how to design such a domain in the case of multiple actions. In the work of~\citealp{Shalit2016}, the domain that is central to the problem of binary treatment estimation is the mixture $\mathcal{D}_T = \mathcal{P}(z)\mathcal{P}(y)$, where $\mathcal{P}(y) = \int_x \rho(y \mid x)d\mathcal{P}(x)$ is the marginal frequency of actions under the logging policy $\rho$. Let us note that in this target domain, the treatment assignment $y$ is randomized. We now show that choosing the same target domain in the setting of multiple treatments allows efficient estimation of the IPM. Indeed, the general problem of computing IPMs is known to be NP-hard. For binary treatments, CFRNet~\cite{Shalit2016} estimated this IPM via maximum mean discrepancy (MMD)~\cite{MMD}. For multiple treatments, recent work~\cite{Atan} focused on the $\mathcal{H}$-divergence~\cite{Ganin2016}. In this work, we propose instead a different nonparametric measure of independence---the Hilbert-Schmidt Independence Criterion (HSIC)~\cite{Gretton2005,hsic}---which also yields an efficient estimation procedure for the IPM.

\begin{prop}
\label{prop:HSIC}
Let us assume that $\mathcal{Z}$ and $\mathcal{Y}$ are separable metric spaces. Let $k: \mathcal{Z} \times \mathcal{Z} \rightarrow \mathbb{R}$ (resp. $l: \mathcal{Y} \times \mathcal{Y} \rightarrow \mathbb{R}$) be a continuous, bounded, positive semi-definite kernel. Let $\mathcal{K}$ (resp. $\mathcal{L}$) be the corresponding reproducing kernel Hilbert space (RKHS). Let us assume that the function space $\mathcal{H}$ is included in a ball of radius $\kappa$ of the tensor space $\mathcal{K} \otimes \mathcal{L}$. Then, one can further bound the IPM in Eq.~\eqref{IPM} as follows:
\begin{equation}
\begin{split}
    \textrm{IPM}_\mathcal{H}\left(\mathcal{D}_S, \mathcal{D}_T\right)
      &\leq \kappa \textrm{HSIC}\left(\mathcal{P}(z)\rho(y \mid \Lambda^{-1}(z))\right).
\end{split}
\label{HSIC_prop}
\end{equation}
\end{prop}
\begin{proof}
See Section 2.3 of~\citealp{smola2007hilbert}, Definition 2 from~\citealp{MMD} and Definition 11 from~\citealp{sejdinovic2013}. 
\end{proof}
Intuitively, the HSIC is a norm of the covariance operator for the joint historical data distribution $(x, z)$ and measures the dependence between the observed action $y$ under the logging policy and $z$. 
Remarkably, the HSIC reduces to the MMD for a binary set of actions (Proposition 2 of \citealp{NIPS2018_7850}) and our theory extends~\citealp{Shalit2016} for multiple as well as continuous actions sets~$\mathcal{Y}$. We provide more context about the derivation of the HSIC and its theoretical properties in Appendix~\ref{HSIC}. Crucially, the HSIC can be directly estimated via samples from $(x, z)$~\cite{hsic} as
\begin{align}
\begin{split}
\hat{\text{HSIC}}_n = &\frac{1}{n^2}\sum_{i, j}^n k(z_i, z_j)l(x_i, x_j) \\ & + \frac{1}{n^4}\sum_{i, j, k, l}^n k(z_i, z_j)l(x_k, x_l) \\
&- \frac{2}{n^3}\sum_{i, j, k}^n k(z_i, z_j)l(x_i, x_k),
\end{split}
\end{align}
where $n$ is the number of samples. Such an estimator can be computed in $\mathcal{O}(n^2)$ operations. 



\subsection{Implementation}
By putting together Eq.~\eqref{IPM} and Eq.~\eqref{HSIC_prop}, we optimize a generalization bound for estimating the incentive function $f$. In our implementation, $\Lambda$ and $g_\psi$ are parameterized by neural networks and we minimize the following loss function during the SIRE step
\begin{align}
\begin{split}
        \mathcal{L}(\psi, \Lambda) =  \frac{1}{n}\sum^n_{i=1} & L\left(g_\psi(\Lambda(x_i), y_i), r_i\right) \\ 
        & + \kappa \textrm{HSIC}\left((\Lambda(x_i), y_i)\right).
\end{split}
\end{align} 
Notably, in order to use stochastic optimization, the HSIC is computed on minibatches of datapoints. We use a linear kernel for all experiments. An important detail is that in practice, the parameter $\kappa$ is unknown. We used a cross-validation procedure to select $\kappa$ in our implementation. Although cross-validation is expected to be biased in this setting, our results suggest a reasonable performance in practice. Then, we use the estimates to solve the CCPO problem. According to Proposition~\ref{prop:eq}, this can be solved easily via a search procedure on the Lagrange multipliers. Also, in the case of discrete values for the cost function, the problem can be solved in time $\mathcal{O}(nM)$ via dynamic programming where $M$ is the total budget.

        

%% file: 5_experiments.tex
\section{Experiments}
\label{sec:exp}
In our experiments, we consider only the case of a discrete action space. We compare our method CCPOvSIRE to a simple modification of BanditNet~\cite{Joachims2018} which handles constrained policy optimization (further details in Appendix~\ref{app:CRM}). We also compare to GANITE~\cite{yoon2018ganite} and to ITE procedures such as BART~\cite{hill2011} and CFRNet~\cite{Shalit2016}. Binary treatment estimation procedures were adapted for multiple treatments following~\cite{yoon2018ganite}. All estimation procedures were employed for CCPO via Proposition~\ref{prop:eq}. As an ablation study, we also compare to two modified algorithms: CCPOvIRE, which is a variant of our algorithm which does not exploit the reward structure and a variant with $\kappa = 0$ (no HSIC). To establish statistical significance, we run each experiment $100$ times and report confidence intervals. For our methods, GANITE and CFRNet, we performed a grid search over the hyperparameters as indicated in Appendix~\ref{app:hyperparam}. As a further sensitivity analysis regarding the choice of the parameter $\kappa$, we report the results of our algorithm for a large range of values for all experiments in Appendix~\ref{app:robustness}. We ran our experiments on a machine with a Intel Xeon E5-2697 CPU and a NVIDIA GeForce GTX TITAN X GPU.

\begin{table*}
\centering
\begin{tabular}{l|c|cc}
                    & \multicolumn{3}{c}{\textbf{Fully-simulated data (mean $\pm$ std)}} \\
                    & \textbf{Binary Action}       & \multicolumn{2}{c}{\textbf{Multiple Actions}}       \\
                    & \textbf{PEHE}                & \textbf{RMSE}                 & \textbf{Reward ($m = 3$)}               \\ \hline
\textbf{CCPOvSIRE}               & NA                   & \textbf{0.0970 $\pm$ 0.0038} & \textbf{0.6082 $\pm$ 0.0013}  \\
\textbf{CCPOvSIRE}$^\dagger$ & NA                   & 0.1115 $\pm$ 0.0135 & 0.6069 $\pm$  0.0020 \\
\textbf{CCPOvIRE}                 & \textbf{0.0385 $\pm$ 0.0011} &  0.1059 $\pm$ 0.0165 & 0.6074 $\pm$ 0.0011  \\
\textbf{CCPOvIRE}$^\dagger$  & 0.0404 $\pm$ 0.0011 & 0.1238 $\pm$ 0.0132 & 0.6061 $\pm$ 0.0009  \\ \hline
\textbf{GANITE}              & 0.1077 $\pm$ 0.0086 & 0.2263 $\pm$ 0.0394 & 0.5964 $\pm$ 0.0054  \\
\textbf{BART}                & 0.0674 $\pm$ 0.0026 & 0.2057 $\pm$ 0.0027 & 0.6002 $\pm$ 0.0001  \\
\textbf{CFRNet}              & 0.0447 $\pm$ 0.0004 & 0.1069 $\pm$ 0.0060 & 0.5958 $\pm$ 0.0010  \\ \hline
\textbf{BanditNet}           & NA                   & NA                   & 0.5899 $\pm$ 0.0076 
\end{tabular}
\caption{Performance with simulated datasets. $\dagger$: baselines with no HSIC ($\kappa$ = 0). Bold indicates the method
with the best performance for each dataset. } 
\label{table:simulations}
\end{table*}

\subsection{Fully-simulated data}
Since it is difficult to obtain a realistic dataset meeting all our assumptions and containing full information, for benchmarking purposes we first construct a synthetic dataset 
of the form $(x, y, r)$
, with five discrete actions and monotonic rewards (further details in Appendix~\ref{app:simu}). We also derive a dataset with binary actions for ITE benchmarking.

To train CCPOvSIRE, we used stochastic gradient descent as a first-order stochastic optimizer with a learning rate of 0.01 and a three-layer neural network with 512 neurons for each hidden layer. First, we report results on the task of estimating the reward function (IRE). For the binary treatment dataset, we report the \textit{Precision in Estimation of Heterogeneous Effect} (PEHE)~\cite{Johansson2016}. For the multiple treatments experiments, we report the \textit{Roots of Mean Squared Error} (RMSE)~\cite{yoon2018ganite}. We then apply all the algorithms to policy optimization for a fixed budget and report the expected reward. In particular, we used the dynamic programming algorithm to solve for the optimal action.

Experimental results on the synthetic dataset for a binary treatment and for multiple treatments, as well as the results for policy optimization with a fixed budget, are shown in Table~\ref{table:simulations}. 
In the binary treatments experiments results, we show that our estimation procedure (IRE) yields the best PEHE. Notably, we observe an improvement over the no-HSIC version, which shows that our HSIC regularization improves the estimation of the reward. In the case of the multiple treatments experiments, we see that CCPOvIRE performs similarly than CFRNet, while CCPOvSIRE improves over all the baselines for reward estimation as well as policy optimization. In particular, we outperform BanditNet by a significant margin on this dataset.

\subsection{Simulating Structured Bandit Feedback from Nested Classification}\label{sec:exp_nest}

\begin{figure}[t]
\center
\subfigure{
\begin{minipage}{0.45\columnwidth}
\centering
\includegraphics[width=\columnwidth]{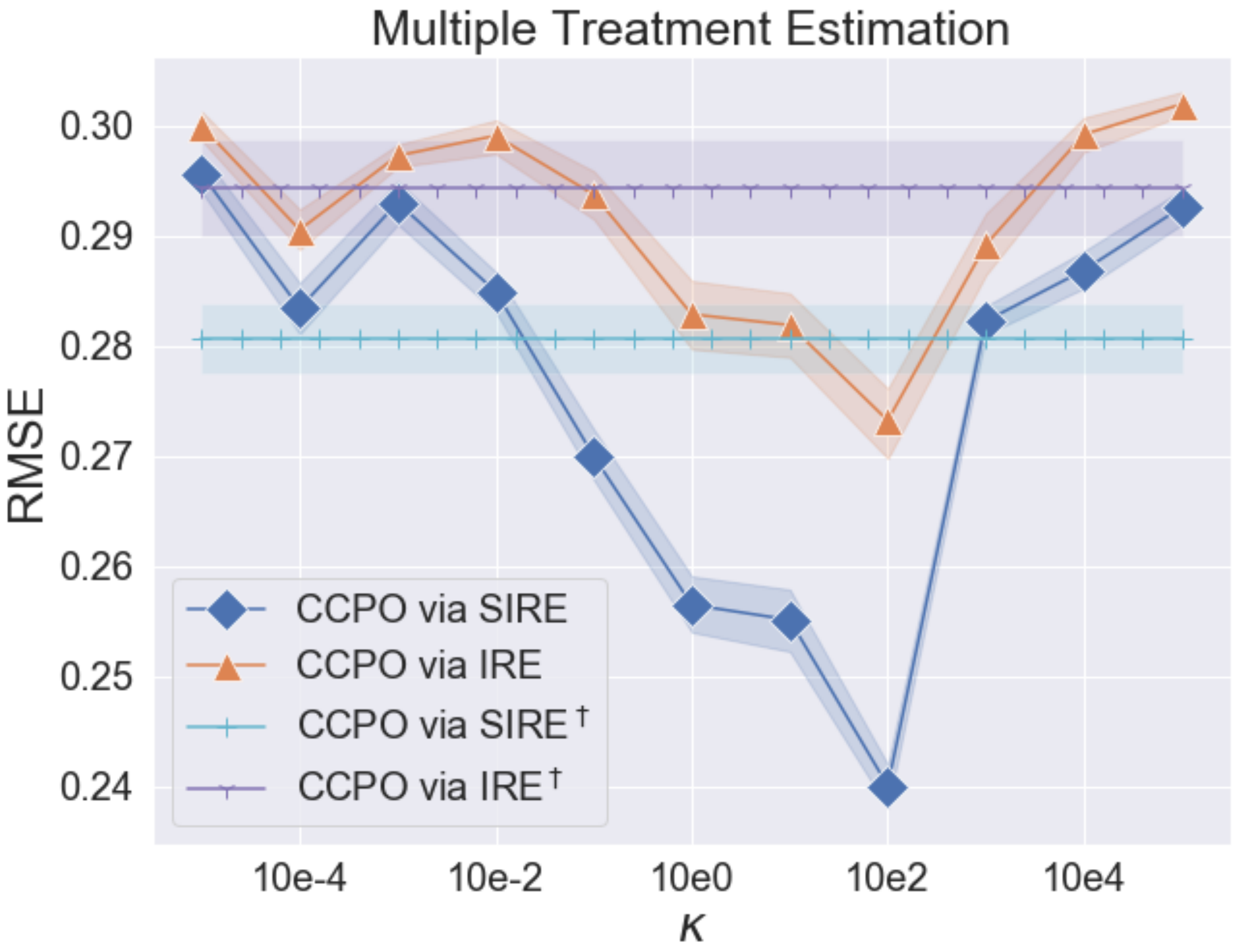}
\footnotesize{(a)}
\end{minipage}
}
\subfigure{
\begin{minipage}{0.45\columnwidth}
\centering
\includegraphics[width=\columnwidth]{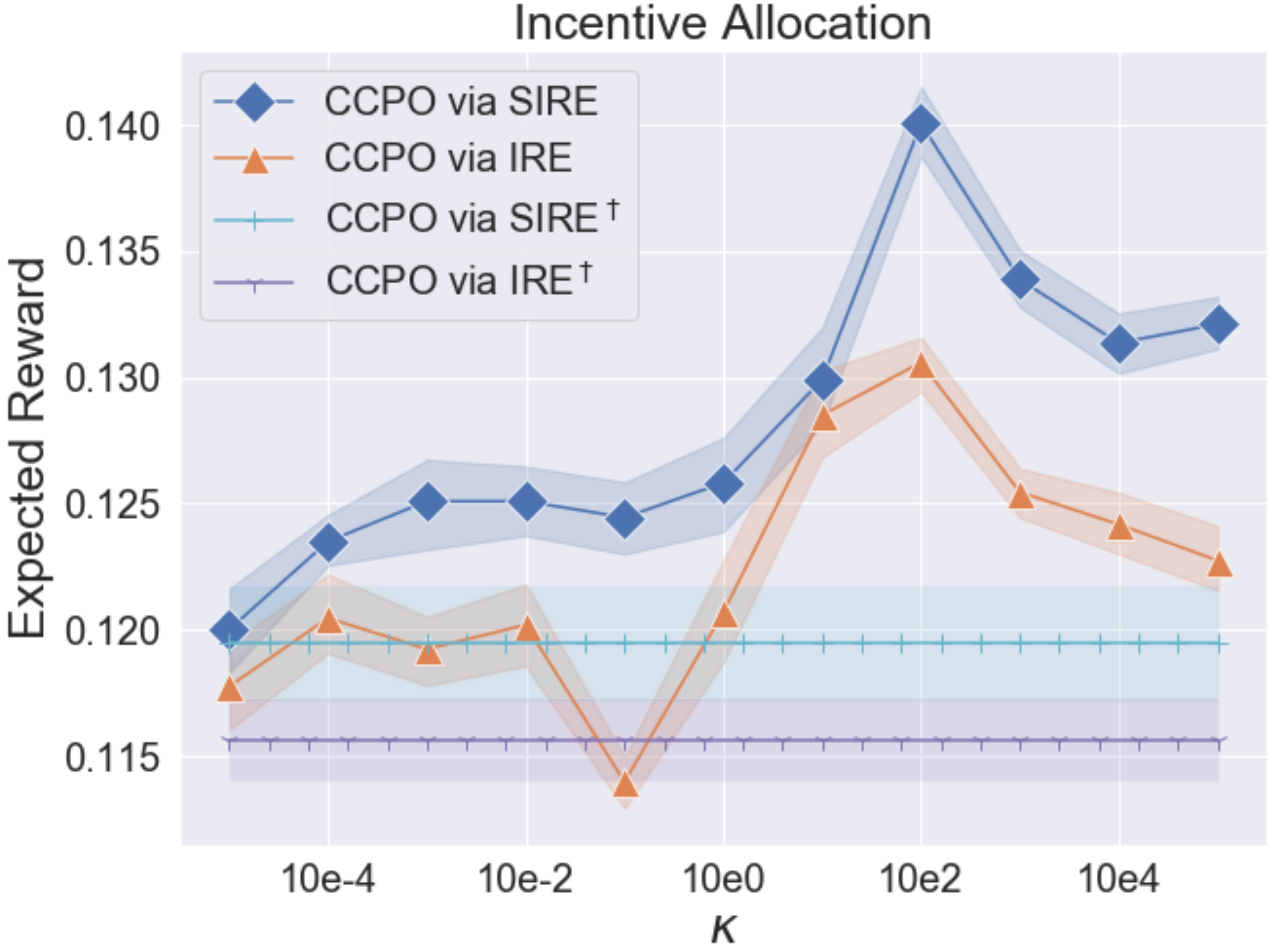}
\footnotesize{(c)}
\end{minipage}
}
\caption{Benchmarking on the ImageNet dataset. Solid lines refer to mean, shadowed areas correspond to one standard deviation. (a) Estimation error with respect to $\kappa$. (b) Policy optimization performance under a budget constraint of $2$ with respect to $\kappa$.}
\label{fig:main}
\end{figure}

Although algorithms for BLFB can be evaluated by simulating bandit feedback in a supervised learning setting~\cite{Agarwal2014}, this approach is not compatible with our structured feedback setting. We therefore propose a novel approach to evaluate cost-effective incentive allocation algorithms. In particular, we use a model of nested classification with bandit-type feedback which brings ordered actions and structured feedback. While the vanilla setting relies on disjoint labels and therefore lacks an ordering between the actions, we simulate nested labels $(\mathcal{Y}_i)_{i \in \mathcal{I}}$, which are therefore monotonic with respect to the set inclusion. We construct a concrete example from the ImageNet dataset~\cite{ImageNet}, by randomly selecting images with labels ``Animal,'' ``Plant,'' and ``Natural object'' and focusing on the nested labels $\textrm{``Animal''} \supset \textrm{``Arthropod''} \supset \textrm{``Invertebrate''} \supset \textrm{``Insect''}$. Let us now derive a compatible reward function. In this particular example of nested classification, say we observe a dataset $(x_i, y_i^*) \sim \mathcal{P}$ where $x_i$ is a pixel value and $y_i^* \in [K]$ is the perfect descriptor for this image among the $K$ labels. Supervised learning aims at finding a labeling function $\Psi^{\textrm{SL}}: \mathcal{X} \rightarrow [K]$ that maximizes $ \E_{(x, y^*) \sim \mathcal{P}}[\mathds{1}_{\Psi^{\textrm{SL}}(x) = y^*}]$. Here, we are interested in a different setting where labels are nested and partial reward should be given when the labels is \emph{correct} but not \emph{optimal}, which corresponds to the reward $\E_{(x, y^*) \sim \mathcal{P}}[\mathds{1}_{\Psi(x) \geq y^*}]$ (after appropriately permuting the labels). Without any additional constraints, the trivial labeling function that returns ``Animal'' yields maximal reward (as in Example~\ref{ex:MV}). By adding a cost function $c$ and budget constraint $m$, the learner will have to guess what is the optimal decision to take (i.e., with an optimal ratio of reward versus cost). Overall, the incentive allocation problem can be written as
\begin{align}
\begin{split}
\max_\Psi ~ &\E_{(x, y^*) \sim \mathcal{P}}[\mathds{1}_{\Psi(x) \geq y^*}] \\
& \textrm{~~such that~~} \E_{(x, y^*) \sim \mathcal{P}} c\left(\Psi(x)\right) \leq m.
\end{split}
\end{align}

\begin{table*}
\centering
\begin{tabular}{l|ccc}
                    & \multicolumn{3}{c}{\textbf{ImageNet semi-simulations (mean $\pm$ std)}}                   \\
                    & \textbf{RMSE}                & \textbf{Reward ($m = 1$)}              & \textbf{Reward ($m = 2$)}              \\ \hline
\textbf{CCPOvSIRE}                & \textbf{0.2408 $\pm$ 0.0183} & \textbf{0.0934 $\pm$ 0.0225} & \textbf{0.1408 $\pm$ 0.0154} \\
\textbf{CCPOvSIRE}$^\dagger$ & 0.2740 $\pm$ 0.0255 & 0.0718 $\pm$ 0.0063 & 0.1186 $\pm$ 0.0198 \\
\textbf{CCPOvIRE}                 & 0.2712 $\pm$ 0.0341 & 0.0839 $\pm$ 0.0137 & 0.1304 $\pm$ 0.0112 \\
\textbf{CCPOvIRE}$^\dagger$  & 0.2943 $\pm$ 0.0345 & 0.0674 $\pm$ 0.0112 & 0.1157 $\pm$ 0.0127\\ \hline
\textbf{GANITE}              & 0.3449 $\pm$ 0.0236 & 0.0679 $\pm$ 0.0217 & 0.0968 $\pm$ 0.0367 \\
\textbf{BART}                & 0.2867 $\pm$ 0.0302 & 0.0492 $\pm$ 0.0217 & 0.0927 $\pm$ 0.0362 \\
\textbf{CFRNet}              & 0.2480 $\pm$ 0.0168 & 0.0861 $\pm$ 0.0220 & 0.1335 $\pm$ 0.0197 \\ \hline
\textbf{BanditNet}           & NA                   & 0.0654 $\pm$ 0.0265 & 0.0997 $\pm$ 0.0367
\end{tabular}
\caption{Performance with the ImageNet dataset. $\dagger$: baselines with no HSIC ($\kappa$ = 0). Bold indicates the method with the best performance for each dataset. } 
\label{table:imagenet}
\end{table*}

A logging policy, with the exact same form as in the fully-simulated experiment, is used to assign one of these four labels for each image. 
The reward function is $1$ if the label is correct, or $0$ otherwise.
The corresponding costs for selecting these labels are $\{3,2,1,0\}$. The dataset has 4608 samples in total, randomly split into training, validation and testing with ratio 0.6: 0.2: 0.2. The ratio of the positive and negative samples is equal to 1:10. Images are uniformly preprocessed, cropped to the same size and embedded into $\mathbb{R}^{2048}$ with a pre-trained convolutional neural network (VGG-16). 
All results are obtained using the same parameters except the number of neurons for the hidden layers that is doubled. In particular, we searched for the optimal Lagrange multiplier in the space $\{1,2,...,10\}$ and returned the largest reward policy within budget constraint.

We report estimation errors as well as expected rewards for two different budgets in Table~\ref{table:imagenet}. 
As we already noticed in the simulations, the bias-correction step via HSIC significantly contributes to improving both the estimation and the policy optimization. Also, we note that after adopting the structured response assumption, both results are improved, which shows the benefit of exploiting the structure. We report results for different values $\kappa$ for treatment estimation and policy optimization with $m = 2$ in Figure~\ref{fig:main}.


We notice that the range of parameters for which CCPOvSIRE improves over ablation studies is large. Furthermore, it is comforting that the same $\kappa$ leads to the smallest estimation error as well as the best performance for policy optimization. Overall, the approach that does not exploit the reward structure (CCPOvIRE) performs similarly to CFRNet and BanditNet for policy optimization. However, CCPOvSIRE, which exploits the structure, outperforms all methods. Finally, it is interesting that the margin between the expected rewards changes for different values of budget. This may be attributable to discrepancies in hardness of detecting the different labels in images, which effectively modulates the difficulty of the incentive allocation depending on the budget.

%

%% file: 6_discussion.tex
\section{Discussion}\label{sec:conclude}

We have presented a novel framework for counterfactual inference based on BLBF scenario but with additional structure on the reward distribution as well as the action space. For this problem setting we have proposed CCPOvSIRE, a novel algorithm based on domain adaptation which effectively trades off prediction power for the rewards against estimation bias. We obtained theoretical bounds which explicitly capture this tradeoff and we presented empirical evaluations that show that our algorithm outperforms state-of-the-art methods based on ITE and CRM approaches. Since the provided plugin approach is not unbiased, further work introducing doubly-robust estimators~\cite{Dudik2011} should lead to better performance.

Our framework involves the use of a nonparametric measure of dependence to debias the estimation of the reward function. Penalizing the HSIC as we do for each mini-batch implies that no information is aggregated during training about the embedding $z$ and how it might be biased with respect to the logging policy. On the one hand, this is positive since we do not have to estimate more parameters, especially if the joint estimation would require solving a minimax problem as in~\cite{yoon2018ganite,Atan}. On the other hand, that approach could be harmful if the HSIC could not be estimated with only a mini-batch. Our experiments show this does not happen in a reasonable set of configurations. Trading a minimax problem for an estimation problem does not come for free. First, there are some computational considerations. The HSIC is computed in quadratic time but linear-time estimators of dependence~\cite{Jitkrittum2016} or random-feature approximations~\cite{Perez-Suay2018} should be used for non-standard batch sizes. 

Following up on our work, a natural question is how to properly choose the optimal $\kappa$, the regularization strength for the HSIC. In this manuscript, such a parameter is chosen with cross-validation via splitting the datasets. However, in a more industrial setting, it is reasonable to expect the central agent to have tried several logging policies which once aggregated into a mixture of deterministic policies enjoy effective exploration properties~(e.g., in \citealp{Strehl2010}). Future work therefore includes the development of a \emph{Counterfactual Cross-Validation}, which would exploit these multiple policies and prevent propensity overfitting compared to vanilla cross-validation.

Another scenario in which our framework could be applied is the case of continuous treatments. That application would be natural in the setting of financial incentive allocation and has already been of interest in recent research~\cite{kallus18a}. The HSIC would still be an adequate tool for quantifying the selection bias since kernels are flexible tools for continuous measurements.

\subsection*{Aknowledgements}
We thank Jake Soloff for useful discussions on nonparametric least squares estimators, which considerably helped framing Proposition 3 of this manuscript.

%% file: supp_material.tex
\clearpage

\appendix

\section{Proof of Proposition 1}
\label{app:proofs}
\begin{proof}
Let us denote the reward function as $f(x, y) = \E[r \mid x, y]$. We use the following estimator for the objective as well as the cost in problem~\ref{CERM}. Let $n \in \mathcal{N}$ and $(x_1, \ldots, x_n) \in \mathcal{X}^n$. An empirical estimate of the expected reward can be written as 
\begin{align}
\begin{split}
\max_{\pi \in \Pi} & \frac{1}{n} \sum_i \sum_y \pi(y \mid x_i) f(x_i, y) \\
& \textrm{~such that~} \frac{1}{n} \sum_i \sum_y c(y)\pi(y \mid x_i) \leq m.
\end{split}
\end{align} 
There exists a Lagrange multiplier $\lambda^*$ such that this problem is has same optimal set than the following
\begin{align}
\begin{split}
\max_{\pi \in \Pi} & \sum_i \sum_y \pi(y \mid x_i) \left(f(x_i, y) - \lambda^* c(y) \right)
\end{split}
\end{align} 
In this particular setting, we can directly solve for the optimal action $y^*_i$ for each customer, which is given by $\argmax_{y \in \mathcal{Y}} f(x_i, y) - \lambda^* c(y)$. To search for the ad-hoc Lagrange multiplier $\lambda^*$, a binary search can be used to make sure the resulting policy saturates the budget constraint.
\end{proof}

\section{Prediction error on non-parametric least-square for asymmetrically regular function classes}
\label{app:nonparam}
As our proposed bound in Proposition~\ref{IPM} holds for $\mathcal{Y}$ discrete or continuous, we will propose a theoretical result in both cases. As these function classes are relatively simple, an open question remains how to extend these results to more complicated structure (e.g., uniformly bounded variation in $x$ and monotonic in $y$ for the discrete case). 

\subsection[Continuous case: smooth function over L2]{Continuous case: $(\alpha, \beta, R)$-smooth functions over $\mathcal{L}^2([0, 1]^2)$}
Let us assume we observe samples $(r_i, [x_i, y_i])_{i=1}^n$ with observation model $r_i = f^*(x_i, y_i) + \sigma w_i$ where $\sigma >0$ and $w_i \sim \textrm{Normal}(0,1)$. Building up on multivariate Fourier series expansions, we can derive an oracle bound on the prediction error for specific function classes (Section 13.3 of~\cite{wainwright_2019}). Namely, we can decompose $f^*$ in $\mathcal{L}^2([0, 1]^2)$ as
\begin{align}
\label{basis}
    \forall (x, y) \in [0, 1]^2, f^*(x, y) = \theta^*_{0} + \sum_{(n_1, n_2) \in \mathbb{N}_*^2} \theta^*_{n_1, n_2} e^{2n_1\pi x + 2n_2\pi y}
\end{align}
Let $\mathcal{F}_{(\alpha, \beta)}(R)$ be the function class of $(\alpha, \beta, R)$-smooth functions:
\begin{align}
    \mathcal{F}_{(\alpha, \beta)}(R) = \left\{ f \in \mathcal{L}^2([0, 1]^2) \mid \int_{[0, 1]^2}\left|\frac{\partial^{\alpha + \beta} f}{\partial x^\alpha \partial y^\beta}\right|^2(x, y)dxdy \leq R\right\}
\end{align}
This function class is interesting to us since it is a simpler way of analyzing functions which can have different smoothness conditions along different axis as long as we have $\alpha \neq \beta$. In particular, it is known that in the univariate case the class of 1-smooth function has similar complexity than the class of monotonic functions (same for 2-smooth and convex functions), which further motivates our study of this toy example. Let us define $\mathcal{G}(M, N)$ the smaller function class
\begin{align}
\begin{split}
    \mathcal{G}(M, N) = &\left\{ (x, y) \mapsto \beta_{0} + \sum_{n_1 = 1}^M \sum_{n_2 = 1}^N \beta_{n_1, n_2} e^{2n_1\pi x + 2n_2\pi y} \right. \\
    &\left. \text{~~such that~~} |\beta_{0}|^2 + \sum_{n_1 = 1}^M \sum_{n_2 = 1}^N |\beta_{n_1, n_2}|^2 \leq 1\right\}
\end{split}
\end{align}
Given that the basis presented in Eq.~\ref{basis} is orthonormal, it is the oracle bound (13.45) in \cite{wainwright_2019} applies and we have for the least-square estimate $\hat{f}$ of $f^*$ over $\mathcal{G}(M, N)$
\begin{align}
    \mathbb{E}_{x, y, w} \left[\frac{1}{n}\sum_{i=1}^n |\hat{f}(x_i, y_i) - f^*(x_i, y_i)|^2\right] \leq \sum_{n_1 = M+1}^\infty \sum_{n_2 = N+1}^\infty |\theta^*_{n_1, n_2}|^2 + \sigma^2 \frac{NM + 1}{n}
\end{align}
Then, assuming that $f^* \in \mathcal{F}_{(\alpha, \beta)}(R)$ and that $f^*$ is of unit norm in $\mathcal{L}^2([0, 1]^2)$, simple properties of Fourier series with boundary conditions implies that the Fourier coefficient will have a polynomial decay $\exists c > 0, \forall (n_1, n_2) \in \mathbb{N}^2,  |\theta^*_{n_1, n_2}| \leq \nicefrac{c}{n_1^\alpha n_2^\beta}$. We therefore have that 
\begin{align}
    \mathbb{E}_{x, y, w} \left[\frac{1}{n}\sum_{i=1}^n |\hat{f}(x_i, y_i) - f^*(x_i, y_i)|^2\right] \leq \frac{cR}{M^{2\alpha}N^{2\beta}} + \sigma^2 \frac{NM + 1}{n}
\end{align}
By choosing $M = N$ proportional to $\left(\frac{n}{\sigma^2}\right)^{\frac{1}{2(\alpha + \beta) + 2}}$, we get the prediction bound
\begin{align}
\mathbb{E}_{x, y, w} \left[\frac{1}{n}\sum_{i=1}^n |\hat{f}(x_i, y_i) - f^*(x_i, y_i)|^2\right] \leq c\left(\frac{\sigma^2}{n}\right)^{\frac{\alpha + \beta}{\alpha + \beta + 1}}
\end{align}
More specifically, we get the following rates:
\begin{itemize}
    \item $\alpha = 1, \beta = 0$ (analogous to smooth in $x$, no assumptions in $y$), slow rate: $\mathcal{O}(n^{-\frac{1}{2}})$
    \item $\alpha = 1, \beta = 1$ (analogous to smooth in $x$, monotonic in $y$), improved rate: $\mathcal{O}(n^{-\frac{2}{3}})$
    \item $\alpha = 1, \beta = 2$ (analogous to smooth in $x$, convex in $y$), improved rate: $\mathcal{O}(n^{-\frac{3}{4}})$
\end{itemize}
These analogies are not exact but are insightful with respect to the possible improvement by assuming more partial structure inside a function class.

\subsection[Discrete case: constant functions]{Discrete case: constant functions over $[0, 1]\times \{1, \ldots, K\}$}
In this setting, we instead rely on metric entropy to count how much smaller can a function class be after adding monotonicity constraints on a discrete covariate. Metric entropy can then be linked to non-asymptotic prediction errors based on chaining argument such as Dudley's entropy principle~\cite{wainwright_2019}. 

\subsubsection{Function spaces and norms}
Let us define the following function classes $\mathcal{F}$ and $\mathcal{F}_m$:
\begin{align}
    \mathcal{F} = \left\{f: [0, 1]\times \{1, \ldots, K\} \rightarrow [0, 1] \mid \exists (a_1, \ldots, a_K) \in [0, 1]^K: \forall y : f(., y) = a_y \right\}
\end{align}
\begin{align}
    \mathcal{F}_m = \mathcal{F} \cap \left\{f: [0, 1]\times \{1, \ldots, K\} \rightarrow [0, 1] \mid \forall x \in [0, 1], f(x, .) \text{~increasing} \right\}
\end{align}
Since we consider only constant functions, it is easy to see that $(\mathcal{F}, \norm{.}'_\infty)$ (where $\norm{.}'_\infty$ is the sup function norm) is isometric to $(\mathcal{U}^K, \norm{.}_\infty)$ where $\mathcal{U}^K = [0, 1]^K$ designates the $K$-dimensional unit cube  and $\norm{.}_\infty$ the infinite norm on $\mathbb{R}^K$. Similarly, $(\mathcal{F}, \norm{.}'_\infty)$ is isometric to $(\mathcal{V}^K, \norm{.}_\infty)$ where $\mathcal{V}^K$ designates the set of vectors $x = (x_1, \ldots, x_K)$ of $\mathcal{U}^K$ such that $x_1 \leq x_2 \ldots, x_{K-1} \leq x_K$. 

\subsubsection{Observation model and M-estimation}
Let $\sigma > 0$, $n \in \mathbb{N}^*$. For $f^*$ in $\mathcal{F}$, we assume the noise model 
\begin{align}
    \forall (i, j) \in \{1, \ldots, n\} \times \{1, \ldots, K\}, r_{ij} = f^*(\nicefrac{i}{n}, j) + \sigma w_{ij}, 
\end{align}
with $w_{ij} \sim \text{Normal}(0, 1)$. Such a noise model is simple since it has a fixed design and assumes that we observe simultaneously all of the components of the vector $x$ (constructed from the isometry property). These assumptions will make it easier to compute the relevant metric entropy (which implies bounds on the localized Gaussian complexity). In particular, we are interested in the least-square estimator
\begin{align}
    \hat{f} = \argmin_{f \in \mathcal{F}} \frac{1}{Kn}\sum_{i=1}^n\sum_{j =1}^K(r_{ij} - f(\nicefrac{i}{n}, j))^2
\end{align}
Such an estimator can be constrained to functions only in $\mathcal{F}_m$, which we note $\hat{f}_m$. The rate of convergence of these M-estimators can be analyzed via empirical processes theory, (Theorem 3.2.5 of~\cite{wellner2013weak}, Corollary 13.7 of~\cite{wainwright_2019}). The central mathematical object for deriving these rates is the celebrated entropy integral
\begin{align}
    J_n(\delta, \mathcal{F}) = \frac{1}{\sqrt{Kn}}\int_0^\delta \sqrt{\log \mathcal{N}(\epsilon, \{f \in \mathcal{F}: \norm{f - f^*}_n \leq \delta\}, \norm{.}_n)}d\epsilon
\end{align}
where $\norm{.}_n$ is the empirical semi-norm defined by $\norm{f}^2_n = \nicefrac{1}{Kn}\sum_{i=1}^n\sum_{j =1}^K\left(f(\nicefrac{i}{n}, j)\right)^2$ and $\mathcal{N}$ designates the covering number. Such a semi-norm can be rewritten with our previously defined isometry as $\norm{f}^2_n = \nicefrac{1}{K}\sum_{j=1}^K(x_j)^2$. Remarkably, we have that $\norm{f}_n \leq \norm{x}_\infty$ and can use the modified entropy 
\begin{align}
    \hat{J}_n(\delta, \mathcal{U}^K) = \frac{1}{\sqrt{Kn}}\int_0^\delta \sqrt{\log \mathcal{N}(\epsilon, \{x \in \mathcal{U}^K: \norm{x - x^*}_\infty \leq \delta\}, \norm{.}_\infty)}d\epsilon,
\end{align}
where $x^*$ is the vector representation of $f^*$.  A similar entropy $\hat{J}_n(\delta, \mathcal{V}^K)$ can be written for the set of monotonic vectors $\mathcal{V}^K$.

\subsubsection{Metric entropy bounds}
In the case of $\mathcal{U}^K$, a volume ratio argument would make the calculations trivial. However, we need to go into further details for $\mathcal{V}^K$ and the underlying constructions for derivations are very similar. So we explicitly write the proof for both, starting by $\mathcal{U}^K$. 

\paragraph{Non-monotonic case:} The sup ball of interest $\mathcal{B}_\infty(x^*, \delta, \mathcal{U}^K)$ can be written as a product space:
\begin{align}
    \mathcal{B}_\infty(x^*, \delta, \mathcal{U}^K) = \{x \in \mathcal{U}^K: \norm{x - x^*}_\infty \leq \delta\} = \bigtimes_{j=1}^K[x^*_j - \delta, x^*_j + \delta].
\end{align}
Let $\epsilon > 0$. For $L = 1 + \lfloor \nicefrac{\delta}{\epsilon}
\rfloor$, let us cut each segment $[x^*_j - \delta, x^*_j + \delta]$ in $L$ regular pieces. The delimiting points are noted $(\theta^j_i)_{i \in \{1, \ldots, L\}}$. Let us now consider the set of points 
\begin{align}
    \Theta = \left\{(\theta^1_{i_1}, \ldots, \theta^K_{i_K})\mid (i_1, \ldots, i_K) \in \{1, \ldots, L\}^K \right\}.
\end{align} 
Let $u = (u_1, \ldots, u_K) \in \mathcal{B}_\infty(x^*, \delta, \mathcal{U}^K)$. By taking for each component $j$ the closest point $\theta^j_*$ in $(\theta^j_i)_{i \in \{1, \ldots, L\}}$, we are sure that $|u_j - \theta^j_*| < \epsilon$. In the sup norm, we therefore showed that the set $\Theta$ $\epsilon$-covers the set $\mathcal{B}_\infty(x^*, \delta, \mathcal{U}^K)$. Since there are $L^K$ distinct points in $\Theta$, we can conclude that 
\begin{align}
    \log \mathcal{N}\left(\epsilon, \mathcal{B}_\infty(x^*, \delta, \mathcal{U}^K), \norm{.}_\infty\right) \leq K \log\left(1 + \frac{\delta}{\epsilon}\right).
\end{align} 
We recognize the covering number of a parametric class of functions with $K$ parameters. Intuitively, without the monotonicity constraint one needs to learn all the constant pieces independently. We argue now that the function class becomes smaller in the case of $\mathcal{F}_m$. 

\paragraph{Monotonic case:} In particular, let us consider the set of vectors 
\begin{align}
    \Theta_m = \left\{(\theta^1_{i_1}, \ldots, \theta^K_{i_K})\mid (i_1, \ldots, i_K) \in \{1, \ldots, L\}^K \text{~and~} \theta^1_{i_1} \leq \theta^2_{i_2} \ldots, \theta^{K-1}_{i_{K-1}} \leq \theta^K_{i_K}\right\}
\end{align}
Let us show now that $\Theta_m$ effectively $\epsilon$-covers $\mathcal{B}_\infty(x^*, \delta, \mathcal{V}^K)$. Let $v = (v_1, \ldots, v_K) \in \mathcal{V}^K$. Let us select for each component $k$ the index $n_k$ such that $\theta_{n_k}$ is the closest scalar to $v_k$ such that $\theta_{n_k} \leq v_k$. We then have a vector $(\theta_{n_1}, \ldots, \theta_{n_K})$, which is at most at $\epsilon$ distance of $v$ and is in $\Theta_m$ (since the flooring function is monotonic, so the image vector is still ordered), which concludes our proof. Furthermore, $\Theta_m$ has cardinal of
\begin{align}
    \left|\Theta_m\right| = \binom{L+K}{K}
\end{align}
which implies the result
\begin{align}
    \log \mathcal{N}(\delta, \mathcal{B}_\infty(x^*, \delta, \mathcal{V}^K), \norm{.}'_\infty) \leq \binom{L+K}{K} = \binom{L+K}{L}.
\end{align} 
 
\subsubsection{Convergence rates}
Plugging these results into a Dudley entropy type bound, we get
 
\paragraph{Non-monotonic case:} 
\begin{align}
    \hat{J}_n(\delta, \mathcal{U}^K) = \frac{1}{\sqrt{Kn}}\int_0^\delta \sqrt{K \log\left(1 + \frac{\delta}{\epsilon}\right)}d\epsilon = \frac{\delta}{\sqrt{n}}\int_0^1\sqrt{\log\left(1 + \frac{1}{t}\right)}dt
\end{align}
Since $\psi_n: \delta \mapsto \frac{\delta}{\sqrt{n}}$ is subhomogeneous of degree strictly less than $2$, and that $\delta_n = \nicefrac{\sigma}{\sqrt{n}}$ verifies the critical inequality $\psi_n(\delta_n) \leq \nicefrac{\delta^2}{\sigma}$, by empirical processes theory~\cite{wellner2013weak} we recover the parametric rate $\mathcal{O}(\nicefrac{\sigma^2}{n})$. It should be noted that because our number of observation ($nK$) scales linearly with the dimension of the function space, the subsequent rate is independent of $K$. Otherwise, we would recover the known rate for high-dimensional regression $\mathcal{O}(\nicefrac{\sigma^2K}{\hat{n}})$ with $\hat{n} = Kn$ the number of observations.

\paragraph{Monotonic case:} 
In this case, we will upper bound the binomial coefficients according to the values of $\epsilon$, which is necessary to derive a better rate than the non-monotonic case. For this, we need to split the modified entropy integral and rely on the symmetry of binomial coefficients: 
\begin{align}
    \hat{J}_n(\delta, \mathcal{V}^K) = \frac{1}{\sqrt{Kn}}\int_0^{\frac{\delta}{K}}\sqrt{\left(\log  \binom{L(\delta, \epsilon)+K}{K}\right)}d\epsilon + \frac{1}{\sqrt{Kn}}\int_{\frac{\delta}{K}}^{\delta}\sqrt{\log \left(\binom{L(\delta, \epsilon)+K}{L(\delta, \epsilon)}\right)}d\epsilon
\end{align}
where the cut at $\frac{\delta}{K}$ comes from the condition $K \leq L$. Then, upper bounding the binomial coefficients accordingly gives 
\begin{align}
    \hat{J}_n(\delta, \mathcal{V}^K) \leq \frac{1}{\sqrt{Kn}}\int_0^{\frac{\delta}{K}}\sqrt{K \log \left(e + \frac{e\delta}{K\epsilon}\right)}d\epsilon + \frac{1}{\sqrt{Kn}}\int_{\frac{\delta}{K}}^{\delta}\sqrt{ \frac{\delta}{\epsilon}\log \left(e + \frac{eK\epsilon}{\delta}\right)}d\epsilon
\end{align}
By changing variables $t = \nicefrac{K\epsilon}{\delta}$, we get 
\begin{align}
    \hat{J}_n(\delta, \mathcal{V}^K) \leq \frac{\delta}{K\sqrt{n}}\int_0^1\sqrt{\log \left(e + \frac{e}{t}\right)}dt + \frac{\delta}{K\sqrt{n}}\int_1^K\sqrt{ \frac{\log(e +et)}{t}}dt
\end{align}
Notably, the function $\eta: K \mapsto \int_1^K\sqrt{ \frac{\log(e +et)}{t}}$ has for equivalent at infinity $\eta(K) \underset{K \rightarrow \infty}{\sim} \sqrt{K \log K}$. Numerical simulations show that for $K \in \{1, \ldots 10^5\}, \eta(K) \leq 2\sqrt{K \log K}$. We therefore conjecture the following bound for the integral entropy
\begin{align}
    \hat{J}_n(\delta, \mathcal{V}^K) \leq c\frac{\delta\sqrt{\log K}}{\sqrt{Kn}},
\end{align}
where $c$ is a universal constant. Notably, the bound on integral entropy satisfies the critical inequality for $\delta_n = \nicefrac{\sigma \sqrt{\log K}}{\sqrt{Kn}}$. We therefore conclude that the prediction error for the least-square estimator is bounded by $\mathcal{O}(\nicefrac{\sigma^2\log K}{nK})$. Once related to the full number of observation $\hat{n}$, we recover a logarithmic dependence in $K$ for the rate $\mathcal{O}(\nicefrac{\sigma^2\log K}{\hat{n}})$, which is a significantly better rate in terms of dependence to the number of actions $K$. 

\section{Non-parametric estimates of dependence with kernels}
\label{HSIC}

Let $(\Omega, \mathcal{F}, \mathbb{P})$ be a probability space. Let $\mathcal{X}$ (resp. $\mathcal{Y}$) be a separable metric space. Let $u: \Omega \rightarrow \mathcal{X}$ (resp. $v: \Omega \rightarrow \mathcal{Y}$) be a random variable. Let $k: \mathcal{X} \times \mathcal{X} \rightarrow \mathbb{R}$ (resp. $l: \mathcal{Y} \times \mathcal{Y} \rightarrow \mathbb{R}$) be a continuous, bounded, positive semi-definite kernel. Let $\mathcal{H}$ (resp. $\mathcal{K}$) be the corresponding reproducing kernel Hilbert space (RKHS) and $\phi: \Omega \rightarrow \mathcal{H}$ (resp. $\psi:\Omega \rightarrow \mathcal{K}$) the corresponding feature mapping.

Given this setting, one can embed the distribution $P$ of random variable $u$ into a single point $\mu_P$ of the RKHS $\mathcal{H}$ as follows:
\begin{equation}
\mu_{P} = \int_{\Omega} \phi(u) P(du).
\label{embedding}
\end{equation}
If the kernel $k$ is universal\footnote{A kernel $k$ is universal if $k(x,\cdot )$ is continuous for all $x$ and the RKHS induced by $k$ is dense in $C(\mathcal{X})$. This is true for the Gaussian kernel $(u, u') \mapsto e^{-\gamma || u - u'|| ^2}$ when $\gamma > 0$.}, then the mean embedding operator $P \mapsto \mu_P$ is injective~\cite{NIPS2007_3340}. 

We now introduce a kernel-based estimate of \emph{distance} between two distributions $P$ and $Q$ over the random variable $u$. This approach will be used by one of our baselines for learning invariant representations. Such a distance, defined via the canonical distance between their $\mathcal{H}$-embeddings, is called the maximum mean discrepancy~\cite{MMD} and denoted $\text{MMD}(P, Q)$. 

The joint distribution $P(u, v)$ defined over the product space $\mathcal{X} \times \mathcal{Y}$ can be embedded as a point $\mathcal{C}_{uv}$ in the tensor space $\mathcal{H} \otimes \mathcal{K}$. It can also be interpreted as a linear map $\mathcal{H} \rightarrow \mathcal{K}$:
\begin{align}
\forall (f, g) \in \mathcal{H} \times \mathcal{K},~ \mathbb{E}f(u)g(v) = \langle f(u) ,\mathcal{C}_{uv}g(v) \rangle_\mathcal{H} = \langle f\otimes g, \mathcal{C}_{uv}\rangle_{\mathcal{H} \otimes \mathcal{K}}.
\end{align}

Suppose the kernels $k$ and $l$ are universal. The largest eigenvalue of the linear operator $\mathcal{C}_{uv}$ is zero if and only if the random variables $u$ and $v$ are marginally independent~\cite{Gretton2005}. A measure of dependence can therefore be derived from the Hilbert-Schmidt norm of the cross-covariance operator $\mathcal{C}_{uv}$ called the Hilbert-Schmidt Independence Criterion (HSIC)~\cite{hsic}. Let $(u_i, v_i)_{1 \leq i \leq n}$ denote a sequence of iid copies of the random variable $(u, v)$. In the case where $\mathcal{X} = \mathbb{R}^p$ and $\mathcal{Y} = \mathbb{R}^q$, the V-statistics in Equation~\ref{hsic_amp} yield a biased empirical estimate~\cite{NIPS2007_3340}, which can be computed in $\mathcal{O}(n^2(p+q))$ time.
An estimator for HSIC is
\begin{align}\label{hsic_amp}
\begin{split}
\hat{\text{HSIC}}_n(P) = \frac{1}{n^2}\sum_{i, j}^n k(u_i, u_j)l(v_i, v_j) &+ \frac{1}{n^4}\sum_{i, j, k, l}^n k(u_i, u_j)l(v_k, v_l) - \frac{2}{n^3}\sum_{i, j, k}^n k(u_i, u_j)l(v_i, v_k).
\end{split}
\end{align}

\section{BanditNet modified baseline}
\label{app:CRM}

We consider again the constrained policy optimization problem in Eq.~\eqref{CERM}. We propose to use as a baseline the self-normalized IPS~\cite{Swaminathan2015self} estimator and a novel algorithmic procedure adapted from BanditNet~\cite{Joachims2018} to solve the constrained policy optimization problem with the stochastic gradients algorithm:
\begin{align}
\begin{split}
\max_\theta \frac{\frac{1}{n} \sum_{i} \delta_i \frac{\pi_\theta(y_i \mid x_i)}{\hat{\rho}(y_i \mid x_i)}}{\frac{1}{n} \sum_{i} \frac{\pi_\theta(y_i \mid x_i)}{\hat{\rho}(y_i \mid x_i)}}
\textrm{~such that~} \frac{1}{n} \sum_{i} \sum_y c(y) \pi_\theta(y \mid x_i) \leq k
\end{split}
\end{align}
Let us assume for simplicity that there exists a unique solution $\theta^*(k)$ to this problem for a fixed budget $k$. This solution has a specific value $S^*(k)$ for the normalization covariate 
\begin{align}
    \frac{1}{n} \sum_{i} \frac{\pi_{\theta^*(k)}(y_i \mid x_i)}{\hat{\rho}(y_i \mid x_i)}.
\end{align}
If we knew the covariate $S^*(k)$, then we would be able to solve for:
\begin{align}
\begin{split}
\max_\theta & \frac{1}{n} \sum_{i} \delta_i \frac{\pi_\theta(y_i \mid x_i)}{\hat{\rho}(y_i \mid x_i)}  \textrm{~such that~} \frac{1}{n} \sum_{i} \frac{\pi_\theta(y_i \mid x_i)}{\hat{\rho}(y_i \mid x_i)} = S^*(k) \textrm{~and~} \frac{1}{n} \sum_{i} \sum_y c(y) \pi_\theta(y \mid x_i) \leq k
\end{split}
\end{align}
The problem, as explained in ~\cite{Joachims2018}, is that we do not know $S^*(k)$ beforehand but we know it is supposed to concentrate around its expectation. We can therefore search for an approximate $S^*(k)$, for each $k$, in the grid $\mathcal{S} = \{ S_1, \ldots, S_p\}$ given by standard non-asymptotic bounds. Also, by properties of Lagrangian optimization, we can turn the constrained problem into an unconstrained one and keep monotonic properties between the Lagrange multipliers and the search of the covariate $S^*(k)$ (see Algorithm~\ref{alg:CCPO}). 

\begin{algorithm}[ht]
  \caption{Counterfactual Risk Minimization baseline
    \label{alg:CCPO}}
\begin{algorithmic}[1]
\STATE Load data in the form $(x, y, r)$
\STATE Fix a budget $k \in \mathbb{R^+}$
\STATE Approximate the logging policy $\hat{\rho}( y \mid x) \approx \rho(y \mid x)$ using a logistic regression or mixture density networks.
\STATE Fix a list of Lagrange multipliers $(\lambda_1, \ldots, \lambda_p) \in \mathbb{R}^p$.
\FOR{$j \in [p]$} \do

	\STATE Solve the following optimization problem with stochastic optimization for a fixed $\eta$ :
    \begin{equation}
    \begin{split}
    \theta_j(\eta) = \argmax_\theta \frac{1}{n} \sum_{i} \bigg[ & (\delta_i - \lambda_j) \frac{\pi_\theta(y_i \mid x_i)}{\hat{\rho}(y_i \mid x_i)}  - \eta \sum_y c(y) \pi_\theta(y \mid x_i) \bigg] 
    \end{split}
    \end{equation}
    \STATE Use binary search on $\eta$ to obtain the highest-cost-but-in-budget policy noted $\theta_j$.
    \STATE Get the $S_j$ corresponding to $\lambda_j$ via 
    $$\frac{1}{n} \sum_{i} \frac{\pi_{\theta_j}(y_i \mid x_i)}{\hat{\rho}(y_i \mid x_i)} = S_j$$
    
\ENDFOR

\STATE Examine the convex envelop for $S$ and make sure it overlaps enough with the Chernoff announced bound (see~\cite{Joachims2018}). Otherwise refine the set of $\lambda$ and rerun the procedure.
\STATE Select the adequate Lagrange multiplier $\lambda$ by following the rule:
$$ \theta^*(k), S^*(k) = \argmax_{(\theta_j, S_j)}\frac{\frac{1}{n} \sum_{i} \delta_i \frac{\pi_{\theta_j}(y_i \mid x_i)}{\hat{\rho}(y_i \mid x_i)}}{S_j}$$ 
\STATE \OUTPUT{($\theta^*(k), S^*(k)$)}
\end{algorithmic}
\end{algorithm}

\section{Hyperparameter optimization}
\label{app:hyperparam}
We performed hyperparameter optimization using held-out data for CFRNet and GANITE (it is known this procedure is suboptimal since we do not have access to counterfactual measurements but has been shown to provide useful enough model selection results in~\cite{Shalit2016}). In particular, we report the grids for search in Table~\ref{table:hyper}.For our method, we used a fixed architecture as described in the experimental section. However, we used the same model selection scheme to keep the best performing value of $\kappa$. We reported results for all the grid (Appendix~\ref{app:robustness}).

\begin{table}[ht]
\centering
\begin{tabular}{l|cc}
\textbf{Hyperparameter} & \textbf{Simulation}                         & \textbf{ImageNet experiment}        \\[0.2cm] 
\textbf{CFRNet}         &                                             &                                      \\
p\_alpha          & \multicolumn{2}{c}{{[}1e-5, 1e-4, 1e-3, 1e-2, 1e-1, 1e0, 1e1{]}}       \\                   
\textbf{GANITE}         &                                             &                                      \\
alpha          & \multicolumn{2}{c}{{[}0, 0.1, 0.5, 1, 2, 5, 10{]}}                                 \\
beta          & \multicolumn{2}{c}{{[}0, 0.1, 0.5, 1, 2, 5, 10{]}}                                 \\
\end{tabular}
\vspace{0.3cm}
\caption{Grids used for hyperparameter search.}
\label{table:hyper}
\end{table}

\paragraph{Choice of kernel} We ran the experiments from Table 1 with a stronger kernel (RBF) to comment on the stability. Indeed, the kernel used was a linear one before (a mistake in the original submission). We found that the results are not extremely sensitive to the choice of the kernel although weaker kernels might work slightly better. This brings a small empirical evidence to our point made in the first paragraph of “IPM estimation via kernel measures of dependency”.

\begin{table}[ht]
    \centering
    \begin{tabular}{lcc}
    & \textbf{RMSE} & \textbf{ Reward(m=3)}  \\
\textbf{RBF} & & \\
\textbf{CCPOvSIRE} & 0.1040 $\pm$ 0.0421 & 0.6002 $\pm$ 0.0005 \\
\textbf{CCPOvIRE} & 0.1205 $\pm$ 0.1063 & 0.6000 $\pm$ 0.0004 \\
\textbf{Linear (in the paper)} & & \\
\textbf{CCPOvSIRE} & 0.0970 $\pm$ 0.0038 & 0.6082 $\pm$ 0.0013 \\
\textbf{CCPOvIRE} & 0.1059 $\pm$ 0.0165 & 0.6074 $\pm$ 0.0011 
    \end{tabular}
    \caption{Alternate results for Table 1 using a more complex kernel}
    \label{tab:my_label}
\end{table}

\section{Robustness to the choice of balance parameter}
\label{app:robustness}

\begin{figure*}[ht]
\center
\subfigure{
\begin{minipage}{0.315\textwidth}
\centering
\includegraphics[width=\textwidth]{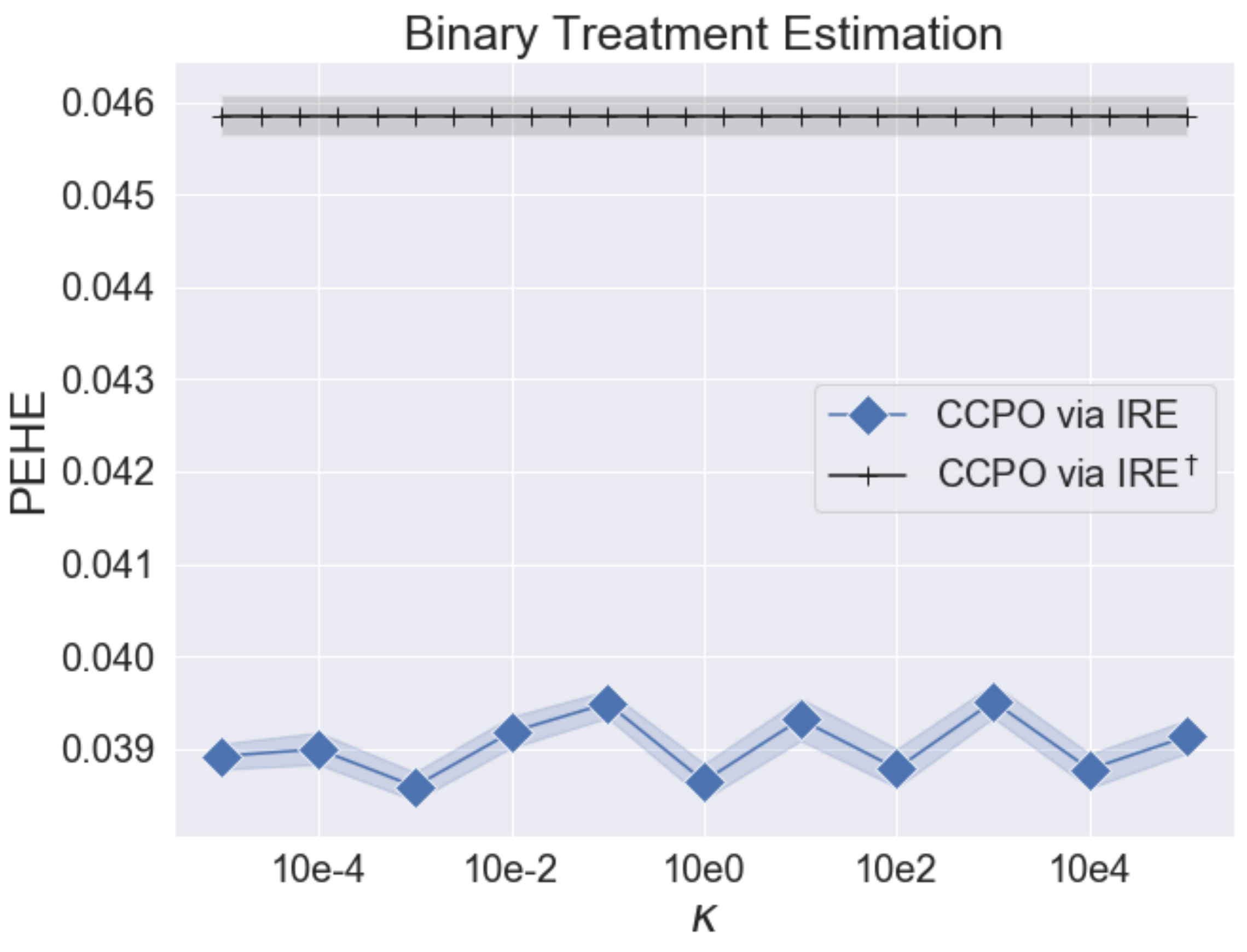}
\label{fig:hsic_curve_binary}
\footnotesize{(a)}
\end{minipage}
}
\subfigure{
\begin{minipage}{0.315\textwidth}
\centering
\includegraphics[width=\textwidth]{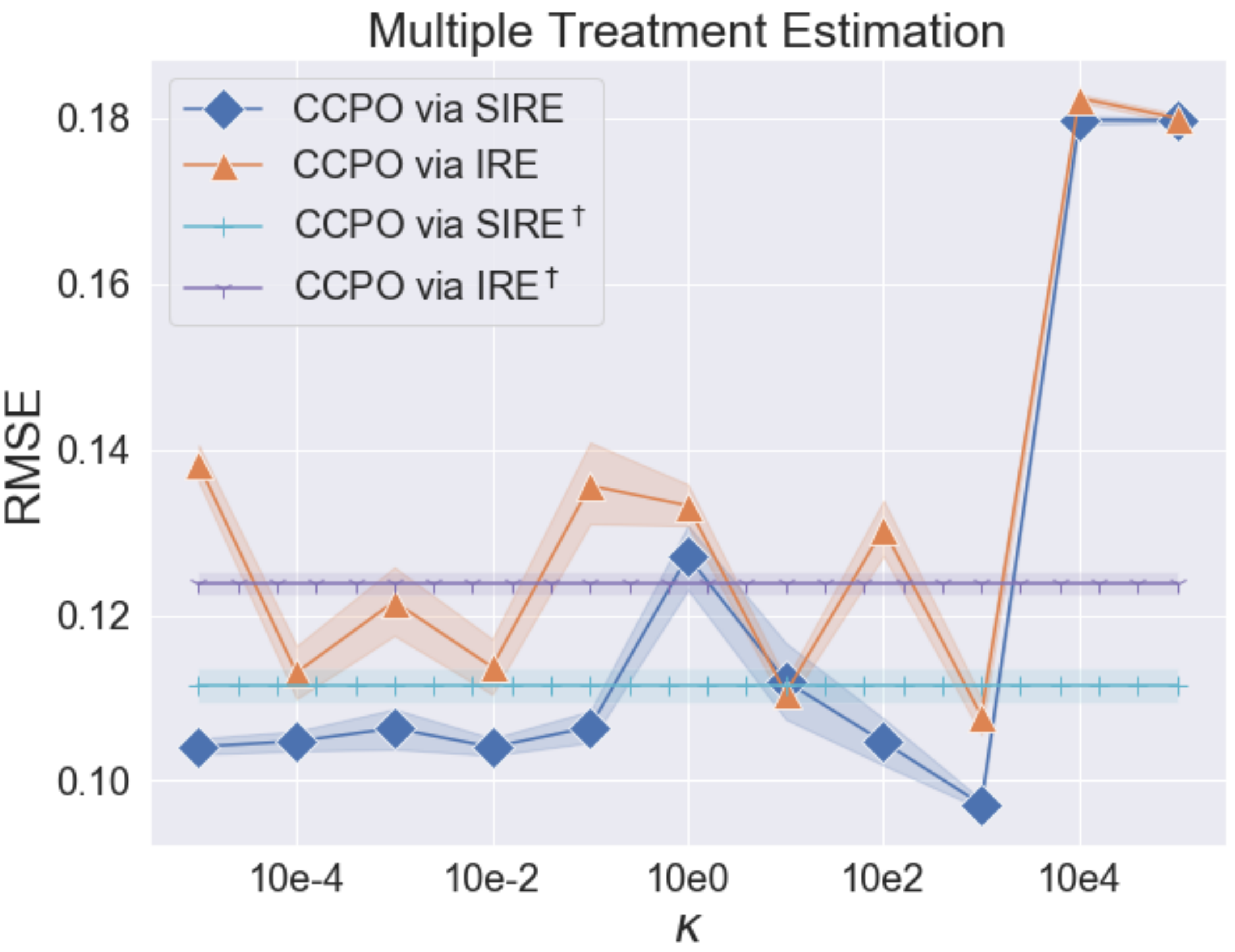}
\label{fig:hsic_curve_multiple}
\footnotesize{(b)}
\end{minipage}
}
\subfigure{
\begin{minipage}{0.315\textwidth}
\centering
\includegraphics[width=\textwidth]{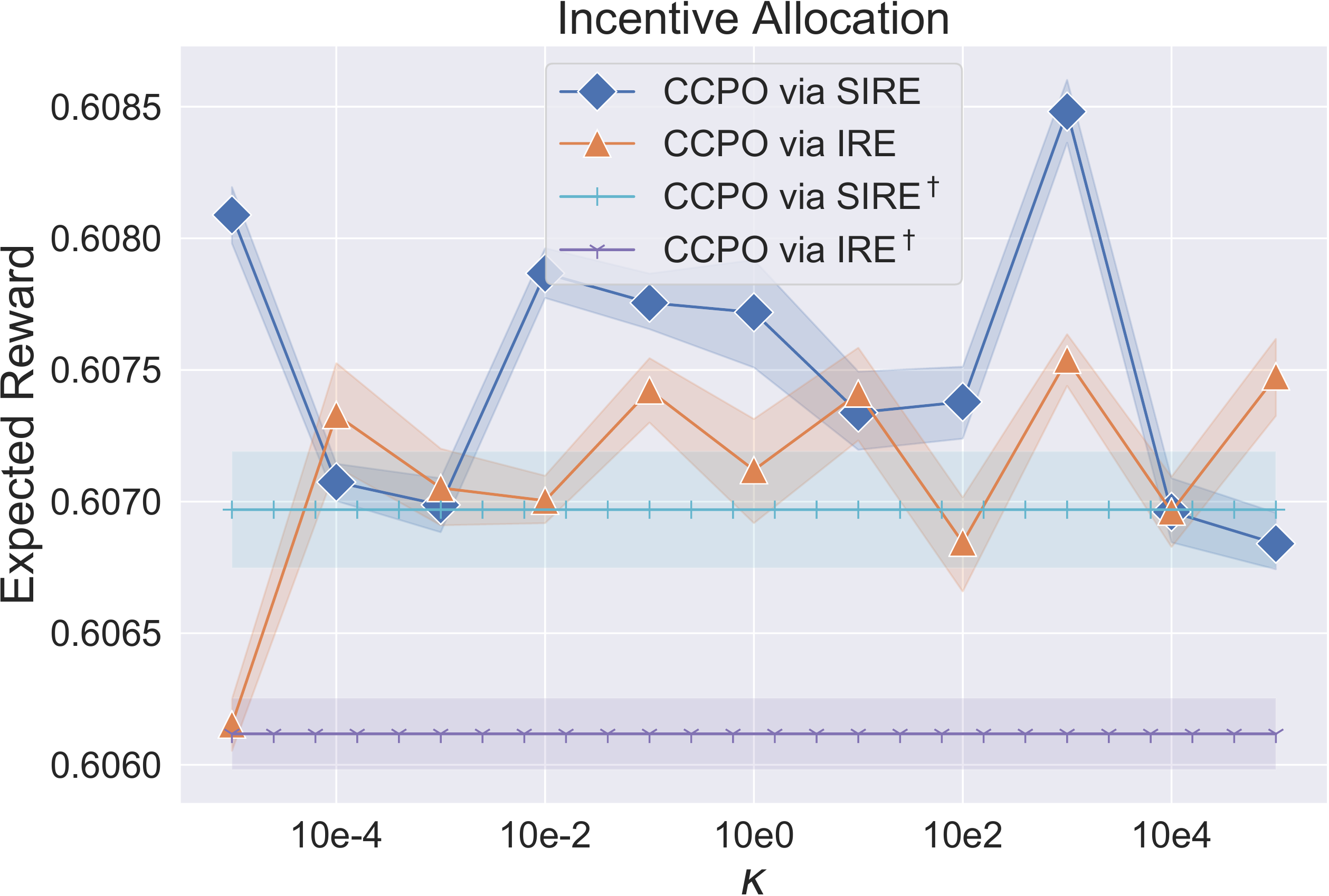}
\label{fig:cpo_simu}
\footnotesize{(c)}
\end{minipage}
}
\caption{Benchmarking on the fully-simulated dataset. Solid lines refer to mean, shadowed areas correspond to one standard deviation. (a) Estimation error with respect to $\kappa$ in the binary treatments experiment. (b) Estimation error with respect to $\kappa$ in the multiple treatments experiment. (c) Policy optimization performance under a budget constraint of $3$ with respect to $\kappa$.}
\label{fig:simu_exp}
\end{figure*}

\begin{figure*}[ht]
\center
\subfigure{
\begin{minipage}{0.315\textwidth}
\centering
\includegraphics[width=\textwidth]{figures/figure2a.pdf}
\label{fig:nested_error}
\footnotesize{(a)}
\end{minipage}
}
\subfigure{
\begin{minipage}{0.315\textwidth}
\centering
\includegraphics[width=\textwidth]{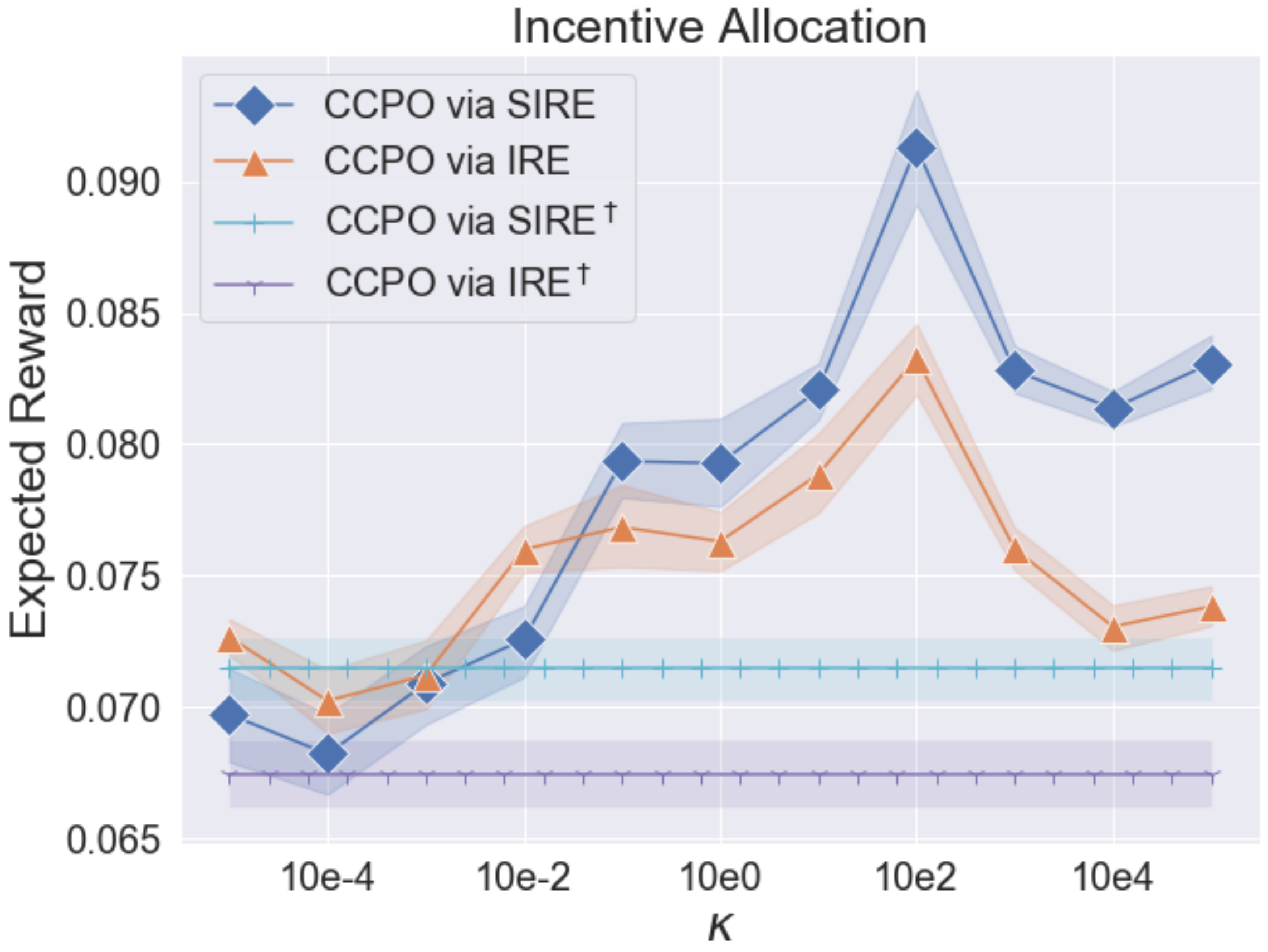}
\label{fig:nested_result}
\footnotesize{(b)}
\end{minipage}
}
\subfigure{
\begin{minipage}{0.315\textwidth}
\centering
\includegraphics[width=\textwidth]{figures/figure2c.pdf}
\label{fig:nested_result_bud1}
\footnotesize{(c)}
\end{minipage}
}
\caption{Benchmarking on the ImageNet dataset. Solid lines refer to mean, shadowed areas correspond to one standard deviation. (a) Estimation error with respect to $\kappa$. (b) Policy optimization performance under a budget constraint of $1$ with respect to $\kappa$. (c) Policy optimization performance under a budget constraint of $2$ with respect to $\kappa$.}
\label{fig:nested_exp}
\end{figure*}

\section{Simulation framework}
\label{app:simu}
Random variable $x$ represents the users' features and is uniformly distributed in the cube $x \sim \mathcal{U}(0, 10)^{50}$. $y \in \{ 1, \ldots, 5 \}$ represents the action historically assigned to the customer, the cost of which is $\{0, \ldots, 4\}$ respectively. Our logging policy $\rho$ is defined as 
\begin{align}
    \rho(y = i \mid x) = \frac{x_i}{\sum_{j=1}^{5} x_j},
\end{align}
where $r$ represents the response after customer $x$ receives the incentive $y$. In order to meet Assumption~\ref{ass:struct}, we generate the response by function $f$ as 
\begin{align}
    f: (x, y) \mapsto S\left(\frac{h(x) - \mu}{\sigma} + \frac{y}{5} \right),
\end{align}
where $\mu$ and $\sigma$ are the mean value and standard deviation of $h(x)$ respectively,  $S$ denotes the sigmoid function and $h$ is defined as
\begin{align}
    h(x_1, \dots, x_{50}) = \sum\limits_{i=1}^{50} a^i \cdot \mathrm{exp} \{ \sum\limits_{j=1}^{50} -b_{j}^{i}|x_j - c_j^i|\},
\end{align}
with $a^i$, $b_j^i$, $c_j^i$ $\in [0, 1]$.
Specifically, we select a group of $a^i$, $b_j^i$ and $c_j^i$ for $i,j \in\{1,2,\dots,50\}$ independently according to the uniform distribution on $[0, 1]$ for each repeated experiment. Each experiment yield a dataset with 500 samples per action, for a total of 2,500 samples. Similarly, we also constructed a binary treatment dataset (only the first two actions are considered as treatments).
